\pdfoutput=1
\documentclass[letterpaper]{article}
\usepackage{spconf,amsmath}
\usepackage[pdftex]{graphicx}

\graphicspath{{./Fig/}}
\DeclareGraphicsExtensions{.pdf,.jpg,.png}

\usepackage[utf8]{inputenc}
\usepackage[T1]{fontenc}

\usepackage{color} 

\usepackage{multirow}

\usepackage{amsfonts}
\usepackage{amsmath,bm}
\usepackage{amsthm} 

\usepackage{mathrsfs}
\usepackage{scalerel}

\usepackage{url} 

\usepackage{etoolbox}
\newcommand{\zerodisplayskips}{%
  \setlength{\abovedisplayskip}{1pt}%
  \setlength{\belowdisplayskip}{1pt}%
  \setlength{\abovedisplayshortskip}{1pt}%
  \setlength{\belowdisplayshortskip}{1pt}}
\appto{\normalsize}{\zerodisplayskips}
\appto{\small}{\zerodisplayskips}
\appto{\footnotesize}{\zerodisplayskips}

\BeforeBeginEnvironment{figure}{\vskip-1ex}
\AfterEndEnvironment{figure}{\vskip-2ex}
\BeforeBeginEnvironment{table}{\vskip-1ex}
\AfterEndEnvironment{table}{\vskip-2ex}

\newcommand\scale[2]{\vstretch{#1}{\hstretch{#1}{#2}}}
	
\newcommand{\LP}{
		\mathbin{\mathpalette\LIPcls+}
	}
\newcommand{\LM}{
		\mathbin{\mathpalette\LIPcls-}
	}

\newcommand{\LIPcls}[2]{%
  \ooalign{$#1\bigtriangleup$\crcr  \hidewidth\raisefix{#1}\hbox{$#1\scale{0.45}{\bm{#2}}$}\hidewidth}}

\def\raisefix#1{%
  \ifx#1\displaystyle
    \raise.14em
  \else
    \ifx#1\textstyle
      \raise.14em
    \else
      \ifx#1\scriptstyle
        \raise.112em
      \else
        \raise.0933em
      \fi
    \fi
  \fi
}

\newcommand{\LIPplus}{\LP}

\newcommand{\Real}{\mathbb R}

\title{Retinal vessel segmentation by probing adaptive to lighting variations}
\name{G. Noyel$^{\star \ddagger}$ \quad C. Vartin$^{\dagger}$ \quad P. Boyle$^{\star \ddagger}$ \quad L. Kodjikian$^{\dagger \ast}$}
\address{$^{\star}$ International Prevention Research Institute, Lyon, France\\
		 $^{\ddagger}$ University of Strathclyde Institute of Global Public Health, Dardilly - Lyon Ouest, France\\
		 $^{\dagger}$ Dpt. of Ophthalmology, Croix-Rousse University Hospital, Hospices Civils de Lyon, Lyon, France\\
		 $^{\ast}$ UMR-CNRS 5510, Mat\'eis, Villeurbane, France}
\begin{document}
%
\maketitle

\begin{abstract}
We introduce a novel method to extract the vessels in eye fundus images which is adaptive to lighting variations. In the Logarithmic Image Processing framework, a 3-segment probe detects the vessels by probing the topographic surface of an image from below. A map of contrasts between the probe and the image allows to detect the vessels by a threshold. In a lowly contrasted image, results show that our method better extract the vessels than another state-of the-art method. In a highly contrasted image database (DRIVE) with a reference, ours has an accuracy of 0.9454 which is similar or better than three state-of-the-art methods and below three others. The three best methods have a higher accuracy than a manual segmentation by another expert. Importantly, our method automatically adapts to the lighting conditions of the image acquisition.
\end{abstract}
%
%
%

\section{Introduction}
\label{sec:intro}

Extracting vessels in eye fundus images has been explored in numerous papers, e.g. \cite{Staal2004,Mendonca2006,Azzopardi2015,Zhao2015,Zhu2017,Hu2018}. 
However, these methods may present limitations when there are strong lighting variations in images. The existence of screening programmes for diabetic retinopathy has led to the creations of large databases of eye fundus images which contain contrast variations. They can be due to: the inhomogeneous absorption of the eye or to different lighting conditions \cite{Noyel2017c}. The aim of this paper is to introduce a vessel segmentation method which is adaptive to these lighting variations in colour eye fundus images. After having complemented the luminance of these images, the vessels appear as a positive relief (i.e. a ``chain of mountains'') in the image topographic surfaces. Their detection is made by a probe composed of three parallel segments, where the central segment has a higher intensity than both others. When the probe is inside a vessel (i.e. a ``mountain''), the intensity difference between its external segments and the bottom of the mountain is minimal, whereas when the probe is outside a vessel, the intensity difference becomes greater. This principle will be used to detect the vessels. The adaptivity to lighting variations is due to the Logarithmic Image Processing model \cite{Jourlin2016}. 
Let us present our method, before showing some experiments and results.

%
%

\section{Method}
\label{sec:meth}


\subsection{Background: Logarithmic Image Processing (LIP)}
\label{ssec:LIP}

Let $f$ be a grey level image defined on a domain $D \subset \Real^n$ with values in $[0,M[ \subset \Real$, where $M$ is equal to 256 for 8-bit digitised images.
The LIP model is based on the \textit{transmittance law} which gives it strong optical properties not only for images acquired by transmission but also by reflection \cite{Jourlin2016}. 
The LIP-addition $\LP$ and its corollary the LIP-subtraction $\LM$ are defined between two images $f$ and $g$ by:
\begin{align}
	f \LIPplus g &= f + g - fg/M, \label{eq:LIP:plus}\\%
	f \LM g 				&= (f-g)/(1-g/M). \label{eq:LIP:minus}%
\end{align}
$f \LM g$ is an image if and only if $f \geq g$. Otherwise, $f\LM g$ may take negative values lying in the interval $]-\infty,M[$. 
In the LIP model, the grey scale is inverted. $0$ corresponds to the white extremity, when no obstacle is placed between the source and the camera. $M$ corresponds to the black extremity, when no light is passing. 
Importantly, the subtraction $\LM$ or the addition $\LP$ of a constant allows to brighten or darken an image as if the light intensity (or the camera exposure-time) was increased or decreased respectively \cite{Jourlin2016}. Such a property allows to adapt to the lighting variations which exist in numerous images.
Let us first present our method in 1D before extending it to 2D.

\begin{figure}[htb]

\begin{minipage}[b]{0.32\linewidth}
  \centering
  \centerline{\includegraphics[width=1\columnwidth]{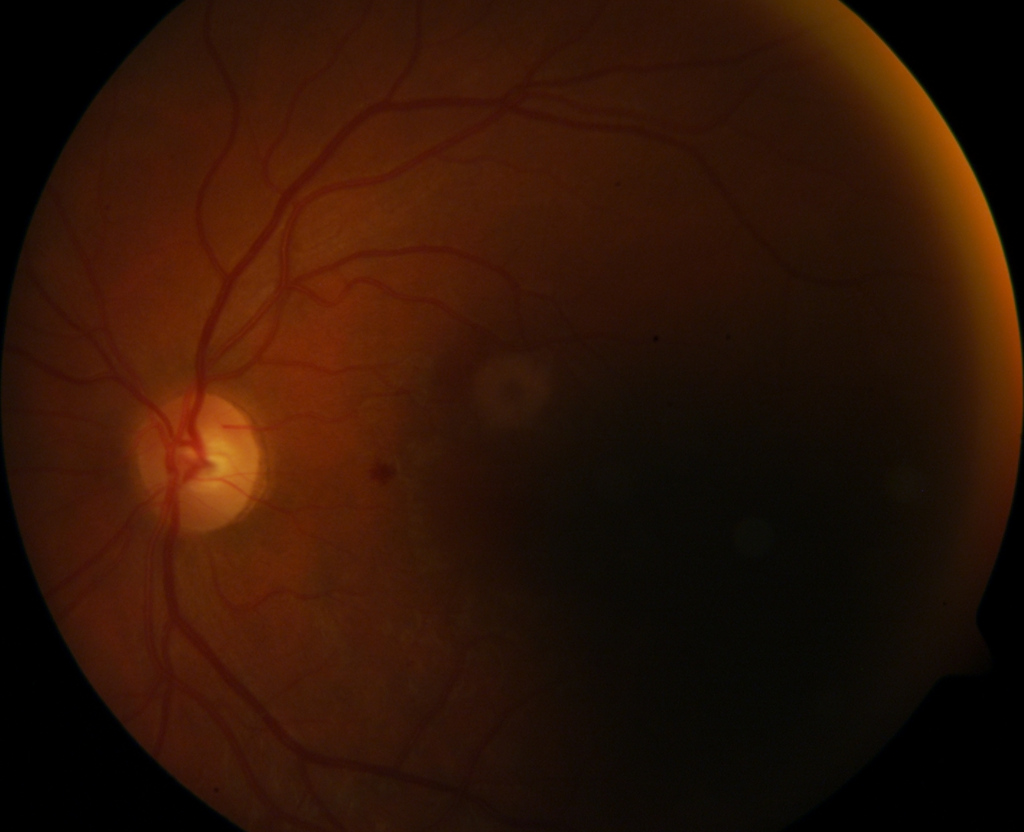}}
  \centerline{(a) Colour image}\medskip
\end{minipage}
\begin{minipage}[b]{0.32\linewidth}
  \centering
  \centerline{\includegraphics[width=1\columnwidth]{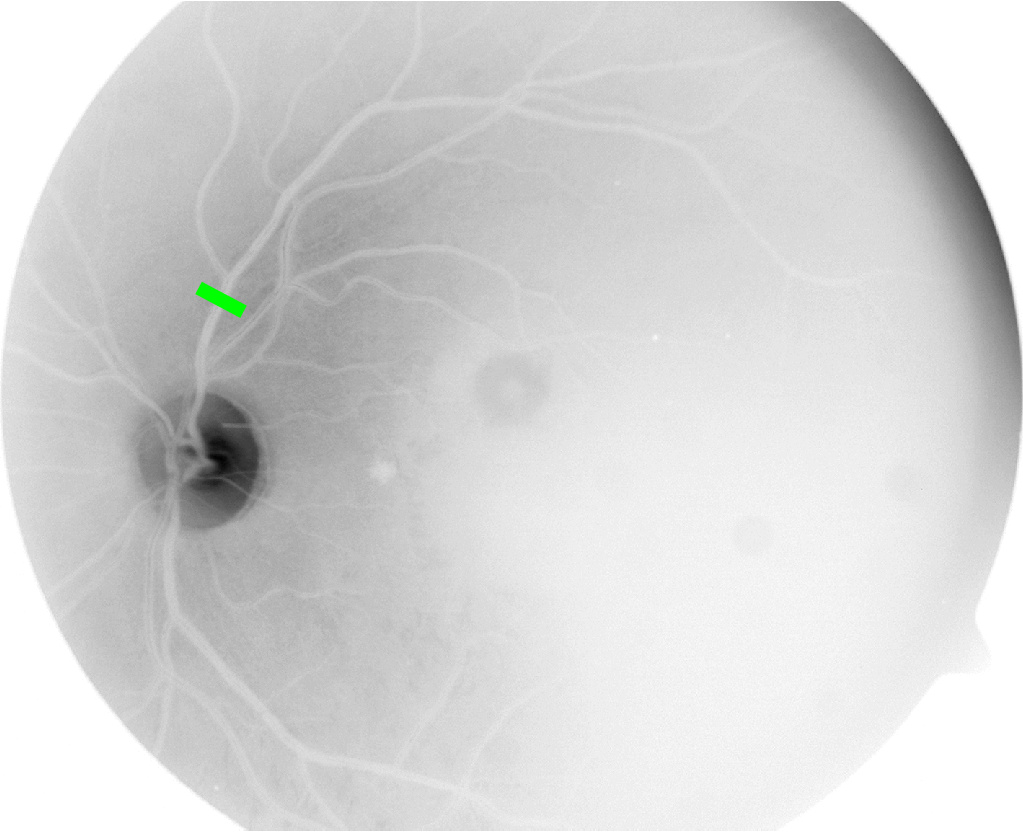}}
  \centerline{(b) $f$}\medskip
\end{minipage}
\begin{minipage}[b]{0.32\linewidth}
  \centering
  \centerline{\includegraphics[width=0.85\columnwidth]{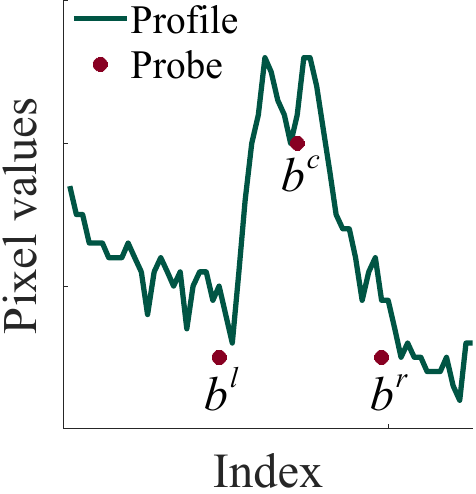}}
  \centerline{(c) Profile and probe}\medskip
\end{minipage}
\caption{(a) A lowly contrasted retinal image \cite{DIARETDB1}. (b) In its luminance image after complementation $f$, a vessel profile is extracted (green segment). (c) Profile $f_p$ and probe $b$.}
\label{fig:profile}
\end{figure}


\subsection{Detection of a vessel profile in 1D}
\label{ssec:detect:1D}

The luminance $\bar{f}$ is equal to $0.299 \times R + 0.587 \times G + 0.114 \times B$, where RGB are the colour components of an image (Fig. \ref{fig:profile} a). To be in the LIP-scale, the luminance is complemented, $f = (M-1)-\bar{f}$ (Fig. \ref{fig:profile} b), and a line-profile $f_p$ of a vessel is extracted (Fig. \ref{fig:profile} c). As the vessel appears as a ``bump'' or a ``mountain'', a probe $b$ made of 3 points is designed to probe this profile from below. The central point of the probe $b^c$ presents a higher intensity than the left one $b^l$ and the right one $b^r$ at the bottom. The distance between the bottom points (i.e. the width of the probe) is larger than the vessel width (Fig. \ref{fig:profile} c). 

\begin{figure}[htb]

\begin{minipage}[b]{0.32\linewidth}
  \centering
  \centerline{\includegraphics[width=1\columnwidth]{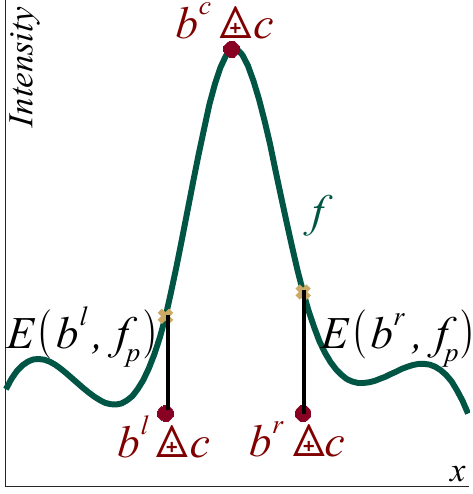}}
  \centerline{(a) $f_1$}\medskip
\end{minipage}
\begin{minipage}[b]{0.32\linewidth}
  \centering
  \centerline{\includegraphics[width=1\columnwidth]{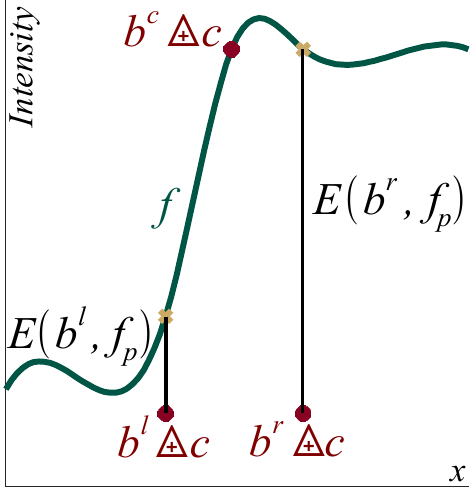}}
  \centerline{(b) $f_2$}\medskip
\end{minipage}
\begin{minipage}[b]{0.32\linewidth}
  \centering
  \centerline{\includegraphics[width=0.5\columnwidth]{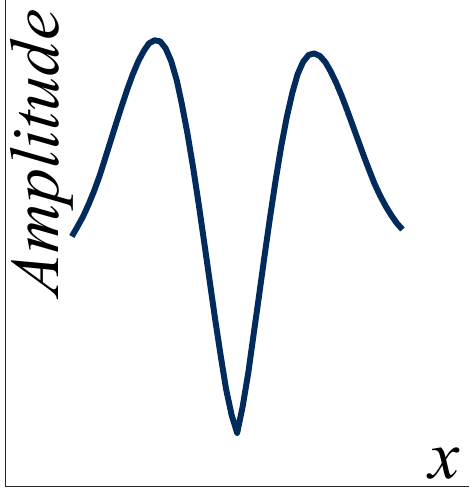}}
  \centerline{(c) $E(b,f_1)$}\medskip
	\centerline{\includegraphics[width=0.5\columnwidth]{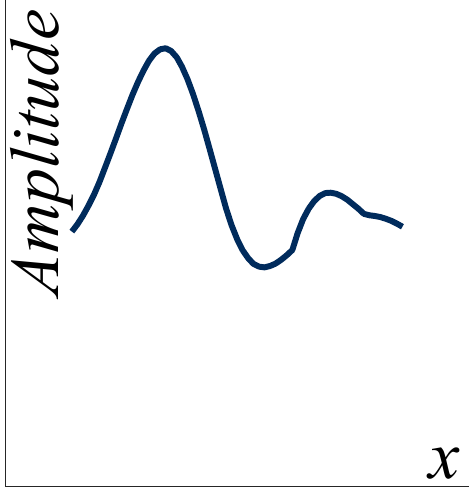}}
  \centerline{(d) $E(b,f_2)$}\medskip
\end{minipage}
\caption{(a) Probing of a bump $f_1$. (b) Probing of a transition $f_2$. The left and right detectors, $E (b^l, f_p)$ and $E (b^r, f_p)$ are shown by vertical black arrows. Maps of the bump detector (c) of the bump $E(b,f_1)$ or (d) of the transition $E(b,f_2)$.}
\label{fig:profile_bump_transi}
\end{figure}

Let us consider the profiles of Fig. \ref{fig:profile_bump_transi} where we want to detect a bump $f_1$ (Fig. \ref{fig:profile_bump_transi} a) but not a transition $f_2$ (Fig. \ref{fig:profile_bump_transi} b). In order to put the 3-point probe $b$ in contact with the profile $f_p$, a constant $c$ is LIP-added to the probe. It is defined by $c = \sup{ \{k, k \LP b \leq f_p \}} = \wedge{ \{ f_p \LM b \}}$ \cite{Jourlin2016}, where $\wedge$ is the infimum. The left and right detectors, $E (b^l, f_p)$ and $E (b^r, f_p)$, are defined as the infimum of the LIP-difference (or contrast \cite{Jourlin2016}) between the profile $f_p$ and the left and right probes, $b^l$ and $b^r$, after the LIP-addition of the constant $c$:
\begin{align}
	E (b^l, f_p) &= \wedge{ \{ f_p \LM [ b^l \LP c ] \} } = \wedge{ \{ f_p \LM b^l \} } \LM c, \label{eq:LAC_er_left}\\
	E (b^r, f_p) &= \wedge{ \{ f_p \LM [ b^r \LP c ] \} } = \wedge{ \{ f_p \LM b^r \} } \LM c. \label{eq:LAC_er_right}
\end{align}
The bump detector is defined as the supremum $\vee$ of the left and right detectors:
\begin{align}
	E (b,f_p) &= \vee{\{ E (b^l, f_p) , E (b^r, f_p) \}}. \label{eq:detector_one_dir}
\end{align}
Maps of bump detectors $E(b,f_{p|D_{b(x)}})$ can be computed by using the restriction of the profile $f_{p}$ to the domain $D_{b(x)}$ of the probe centered in each point $x$. In case of a bump, the map presents a deep minimum (Fig. \ref{fig:profile_bump_transi} c), whereas in case of a transition, this minimum disappears (Fig. \ref{fig:profile_bump_transi} d).


\begin{figure}[htb]
\begin{center}
\begin{minipage}[b]{0.32\linewidth}
  \centering
  \centerline{\includegraphics[width=1\columnwidth]{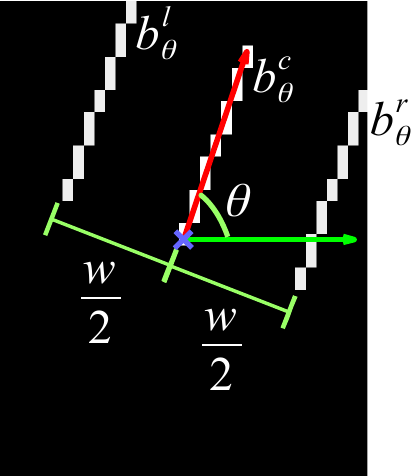}}
  \centerline{(a) Probe domain}\medskip
\end{minipage}
\hfil
\begin{minipage}[b]{0.38\linewidth}
  \centering
  \centerline{\includegraphics[width=1\columnwidth]{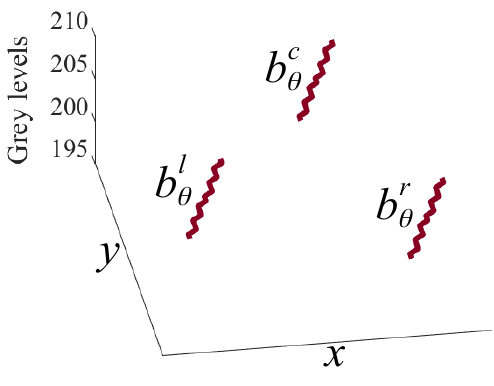}}
  \centerline{(b) Probe intensity}\medskip
\end{minipage}
\end{center}
\caption{(a) 2D probe $b$ with an orientation $\theta$ and a width $w$. (b) The central segment $b^c_{\theta}$ has a higher intensity than both others ones $b^l_{\theta}$ and $b^r_{\theta}$.}
\label{fig:Probe}
\end{figure}


\subsection{Detection of the vessels in 2D}
\label{ssec:detect:2D}

As retinal images are 2D functions, the probe $b_{\theta}$, defined on $D_{b_{\theta}} \subset D$, is made of 3 parallel segments with the same length $l$ and orientation $\theta$ (Fig. \ref{fig:Probe} a). The origin of the probe corresponds to one of the extremities of the central segment $b^c_{\theta}$. Its intensity 
is greater than the intensity 
of the left and right segments $b^l_{\theta}$ and $b^r_{\theta}$ (Fig. \ref{fig:Probe} b). These two segments are equidistant of the central one and the width of the probe is $w$.
In order to define robust to noise operators, we will use the $k^{th}$ minimum, $\wedge^k$, defined as the $k^{th}$ element of a set $X={\{x_1, \ldots x_k, \ldots x_n\}}$ sorted in descending order, $\wedge^k(X)=x_k$.

Given $f$ the complemented luminance of a retinal image, the maps of the left and right detectors, $E_{b^l_{\theta}} f$ and $E_{b^r_{\theta}} f$, are for all $x\in D$ equal to:
\begin{align}
	E_{b^l_{\theta},k} f(x) &= \wedge^k_{ h \in D_{b^l_{\theta}} }{ \{ f(x+h) \LM b^l_{\theta}(h) \} \LM \grave{c}_{\protect b_{\theta},k}f(x) } \label{eq:map_LAC_er_left}\\
	E_{b^r_{\theta},k} f(x) &= \wedge^k_{ h \in D_{b^r_{\theta}} }{ \{ f(x+h) \LM b^r_{\theta}(h) \} \LM \grave{c}_{\protect b_{\theta},k}f(x) } \label{eq:map_LAC_er_right}
\end{align}
The constant map $\grave{c}_{\protect b_{\theta},k}f$ is the point-wise infimum $\wedge$ of the constant maps $c_{b^c_{\theta}}f$, $c_{b^l_{\theta},k}f$ and $c_{b^r_{\theta},k}f$ for each segment $b^c_{\theta}$, $b^l_{\theta}$ and $b^r_{\theta}$: 
$\grave{c}_{b_{\theta},k}(f) = \wedge{\{\>  c_{b^c_{\theta}}f , \>	\wedge{[ c_{b^l_{\theta},k}f , c_{b^r_{\theta},k}f ]} \>\} }$.
 As the central segment $b^c_{\theta}$ must fully enter in the vessel relief, the infimum must be exact and the map $c_{{\protect b^c_{\theta}}} f = \wedge_{h \in b^c_{\theta}} \{ f(x+h) \LM b^c_{\theta}(h) \}$ will be used. However, for the left and right segments, $b^l_{\theta}$ and $b^r_{\theta}$, the robust to noise maps $c_{{\protect b^l_{\theta},k}}f = \wedge^k_{h \in b^l_{\theta}} \{ f(x+h) \LM b^l_{\theta}(h) \}$ and $c_{{\protect b^r_{\theta},k}}f$ will be used. 

The bump detector map in the orientation $\theta$, $E_{b_{\theta},k} f$, is defined as the point-wise supremum $\vee$ of:
\begin{align}
	E_{b_{\theta},k} f &= \vee{\{ E_{b^l_{\theta},k} f , E_{b^r_{\theta},k} f \}}. \label{eq:detector_one_or}
\end{align}
The bump detector map is expressed as the point-wise infimum of the maps $E_{b_{\theta},k} f$ in all the orientations $\theta \in \Theta$:
\begin{align}
	E_{b,k} f &= \wedge{\{ E_{b_{\theta},k} f , \theta \in \Theta \}} \label{eq:detector_tol_1probe}
\end{align}
As the vessel detection is a multi-scale problem, $I$ different probes $\{b_i\}_{i \in [\![1 \ldots I]\!]}$, of widths $\{w_i\}_i$ and length $\{l_i\}_i$ will be used.
The bump detector maps $E_{b_i,k} f$ for the probes $b_i$ are then combined by point-wise infimum:
\begin{align}
	e^I_{b,k} f &= \wedge{\{ E_{b_i,k} f , i \in [\![1 \ldots I]\!] \}} \label{eq:map_detector_tol_3probes}
\end{align}
In the \textit{map of vesselness} $e^I_{b,k} f$ (Fig. \ref{fig:seg_drk_im} a), the vessels appear as valleys and can be segmented by a threshold.

For a better visualisation, the map of vesselness can be normalised as follows. As the vessel values are less than the median $\mu$ of the map $e^I_{b,k} f$ (Fig. \ref{fig:seg_drk_im} b), we define a new map:
$\bar{e}_{b,k} f(x) =	e^I_{b,k} f(x)$, if $e^I_{b,k} f(x) \leq \mu$ or $\bar{e}_{b,k} f(x) = \mu$, else.
The values of the map $\bar{e}_{b,k}$ are set in the interval $[0,1]$ in order to define the normalised map $\Phi_{b,k} f$ (Fig. \ref{fig:seg_drk_im} d), for all $x \in D$, by:
\begin{align}
	\Phi_{b,k} f(x) &= 1 - \frac{\bar{e}_{b,k}(x) - \min\{ \bar{e}_{b,k} \}}{ \max\{ \bar{e}_{b,k} \} - \min\{ \bar{e}_{b,k} \}}.
	 \label{eq:map_norm}
\end{align}

%
%

\section{Experiments and results}
\label{sec:res}


\subsection{Experiments for parameter estimation}
\label{ssec:exp:param}

Experiments were performed in lowly contrasted images from DIARETDB1 database \cite{DIARETDB1} (Fig. \ref{fig:profile}) and in highly contrasted images from DRIVE database \cite{Staal2004}. DIARETDB1 images were captured with a Field Of View (FOV) of $50$ degrees  \cite{DIARETDB1}, whereas in DRIVE the FOV angle was $45^{\circ}$. Parameters are normalised to be the same for all the images. Each parameter is carefully chosen so as to obtain the best segmentation results. A DIARETDB1 image is used for a qualitative evaluation, whereas DRIVE images are used for a quantitative evaluation. Indeed, it contains 20 images with a reference given by an expert. 
The parameters are as follows. The $k^{th}$ minimum, $\wedge^k$, is chosen such that $20 \%$ of the minimal points of a set are discarded. $18$ orientations $\theta$ between $0$ and $360^{\circ}$ were found sufficient. A maximum number of 3 probes $b_1$, $b_2$ and $b_3$ will be used. Their widths are related to the FOV diameter $D_{FOV}$ of the image and the ratio between the FOV angle of a reference camera, $\alpha_r=45^{\circ}$, and the FOV angle of the image camera, $\alpha$. As the width $w_1$ must be greater than the diameter of the largest vessels, $w_1 = (D_{FOV}/50)(\alpha_{r}/\alpha)$ is appropriate. The width $w_2$ and $w_3$ are equal to $w_2 = 0.75 w_1$ and $w_3 = 0.5 w_1$. As the smallest vessels may be more tortuous than the largest ones, the length $l_i$ of a probe must be smaller than its width $w_i$. We will use $l_i = 0.75 w_i$. 
The intensity of the probes will depend on the image mean value $m_f$. Initially, the central segment intensity is set at $h^{c}_{ref} = 215$ and the left and right segment intensities at $h^{r}_{ref}=h^{l}_{ref}=225$. For each image $f$, the central segment intensity is then equal to $h^{c} = m_f$ and the left and right segment intensities to $h^{r}=h^{l}= h^{r}_{ref} \LP (h^{c} \LM h^{c}_{ref})$. The map of vesselness $e^I_{b,k} f$ is segmented with a threshold so that $12 \%$ of the FOV area are considered as vessels (Fig. \ref{fig:seg_drk_im} a). In order to avoid the segmentation of zones of noise, less than $3$ probes may be used. The probe number $I$ is chosen by verifying that the number of pixels whose class is changing between the segmentations of the vesselness maps with $I$ probes $seg(e^I_{b,k} f)$ and with the first probe $seg(e^1_{b,k} f)$ does not exceed $40 \%$ of the vessel area of the segmentation $seg(e^1_{b,k} f)$. The selected segmentation $seg(e^I_{b,k} f)$ is then filtered: the regions whose area is less than $(w_1/2)^2$ are removed and the small holes of the vessels are filled  (i.e the complemented segmentation is eroded by a unit square and reconstructed by dilation under the complemented segmentation). 
Moreover, for the vesselness map $e^I_{b,k} f$ and the normalised map $\Phi_{b,k} f$, only the values which are inside the FOV mask are considered. In DIARETDB1 database, the FOV masks are segmented by a threshold, whereas they are available in DRIVE database.


\subsection{Qualitative results in a lowly contrasted image}
\label{ssec:exp:drk_im}

\begin{figure}[htb]
\begin{center}
\begin{minipage}[b]{0.32\linewidth}
  \centering
  \centerline{\includegraphics[width=1\columnwidth]{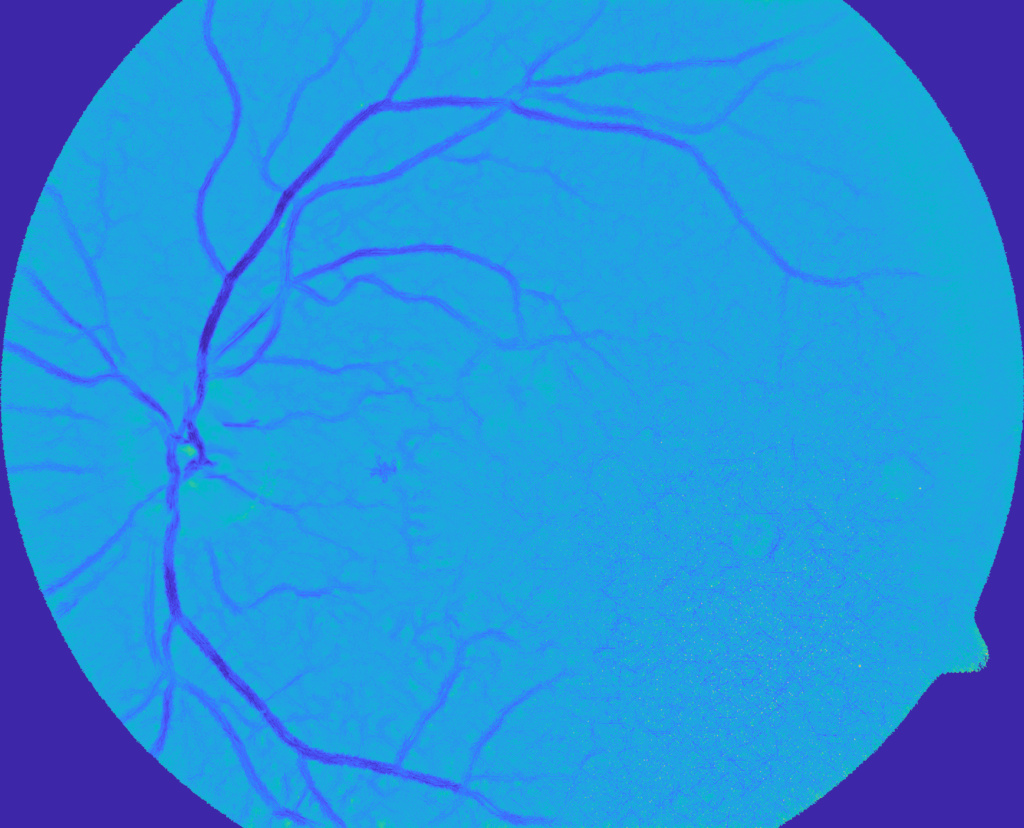}}
  \centerline{(a) $e^I_{b,k} f$}\medskip
\end{minipage}
\begin{minipage}[b]{0.32\linewidth}
  \centering
  \centerline{\includegraphics[width=1\columnwidth]{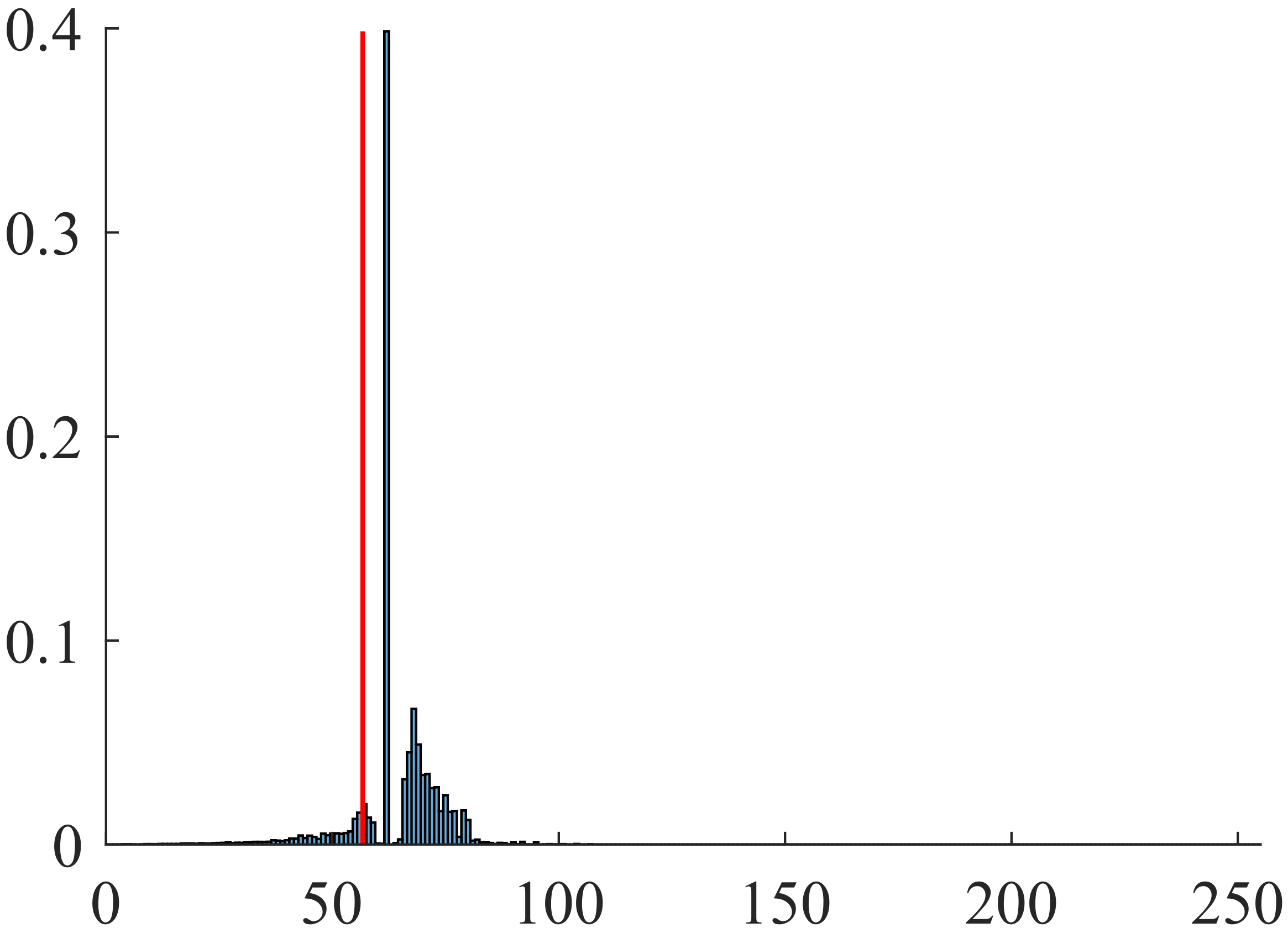}}
  \centerline{(b) Histogram}\medskip
\end{minipage}
\begin{minipage}[b]{0.32\linewidth}
  \centering
  \centerline{\includegraphics[width=1\columnwidth]{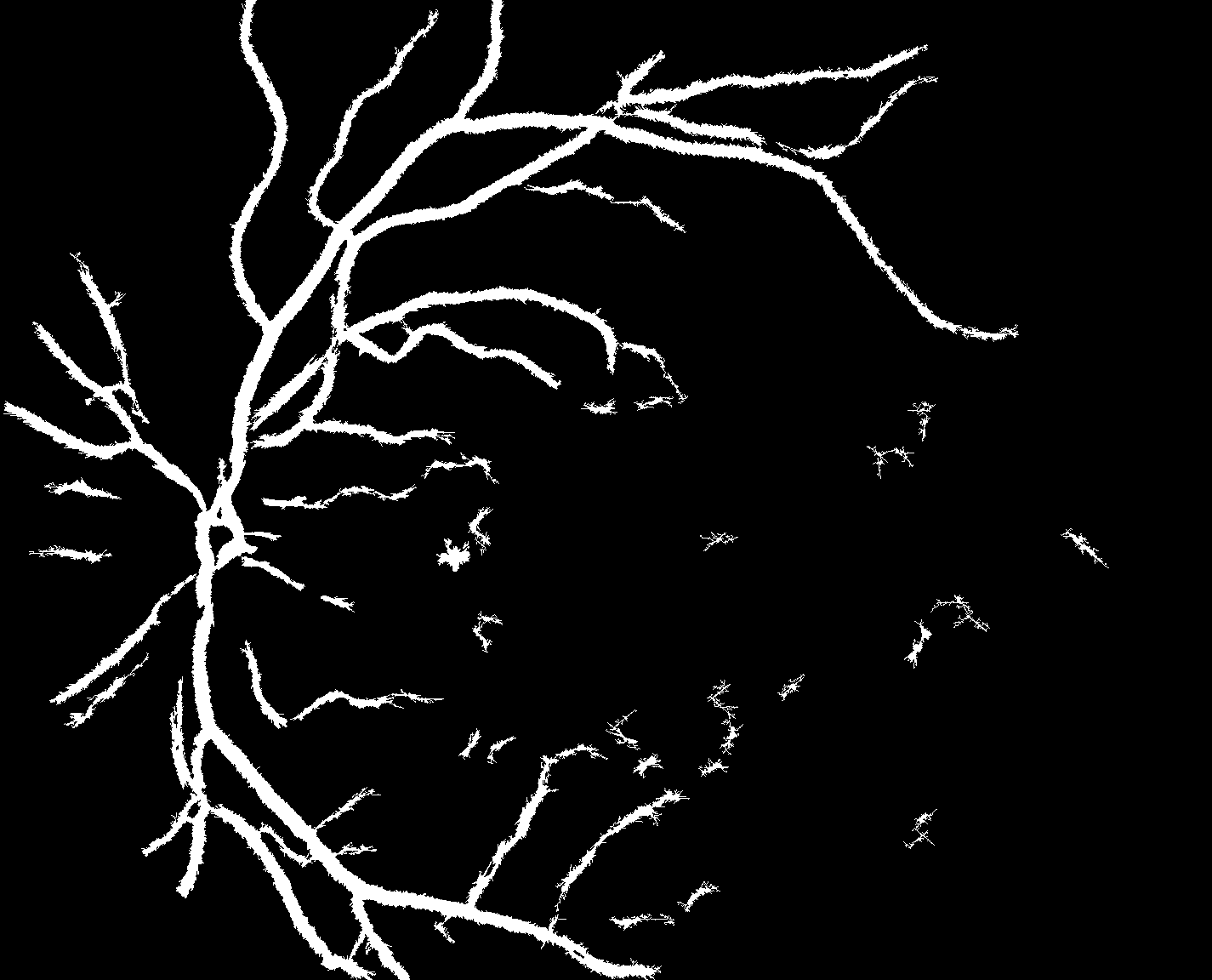}}
  \centerline{(c) Segmentation}\medskip
\end{minipage}\\
\begin{minipage}[b]{0.32\linewidth}
  \centering
  \centerline{\includegraphics[width=1\columnwidth]{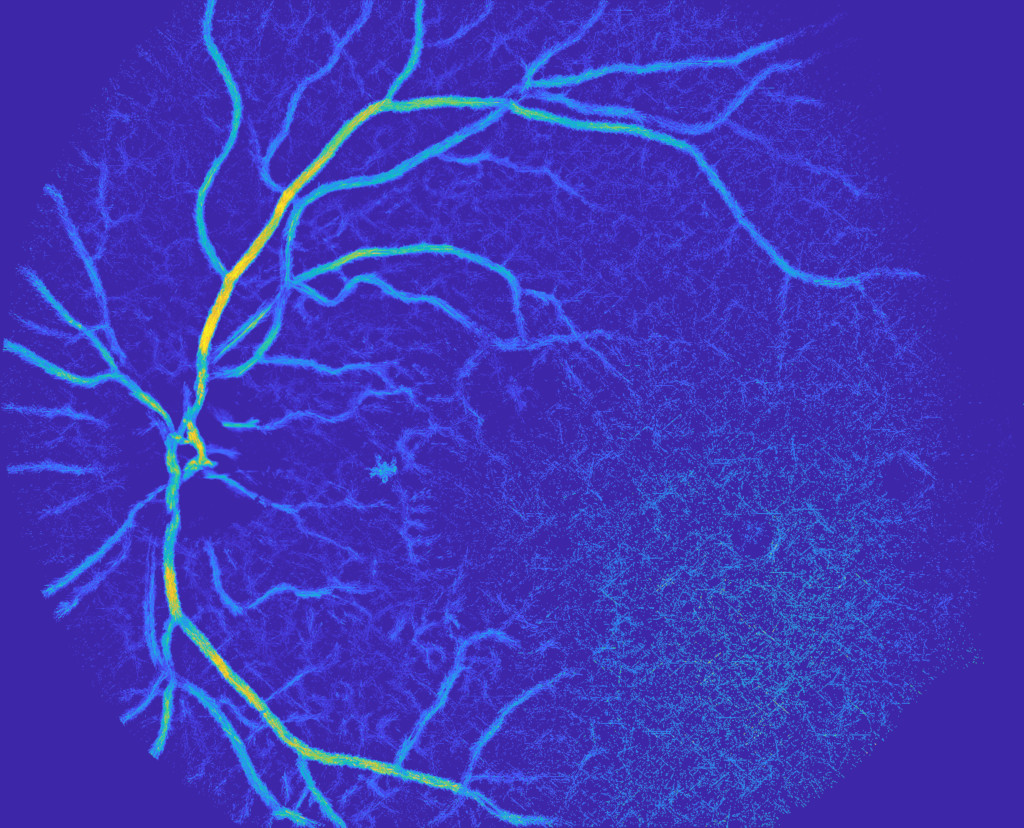}}
  \centerline{(d) $\Phi_{b,k} f$}\medskip
\end{minipage}
\begin{minipage}[b]{0.32\linewidth}
  \centering
  \centerline{\includegraphics[width=1\columnwidth]{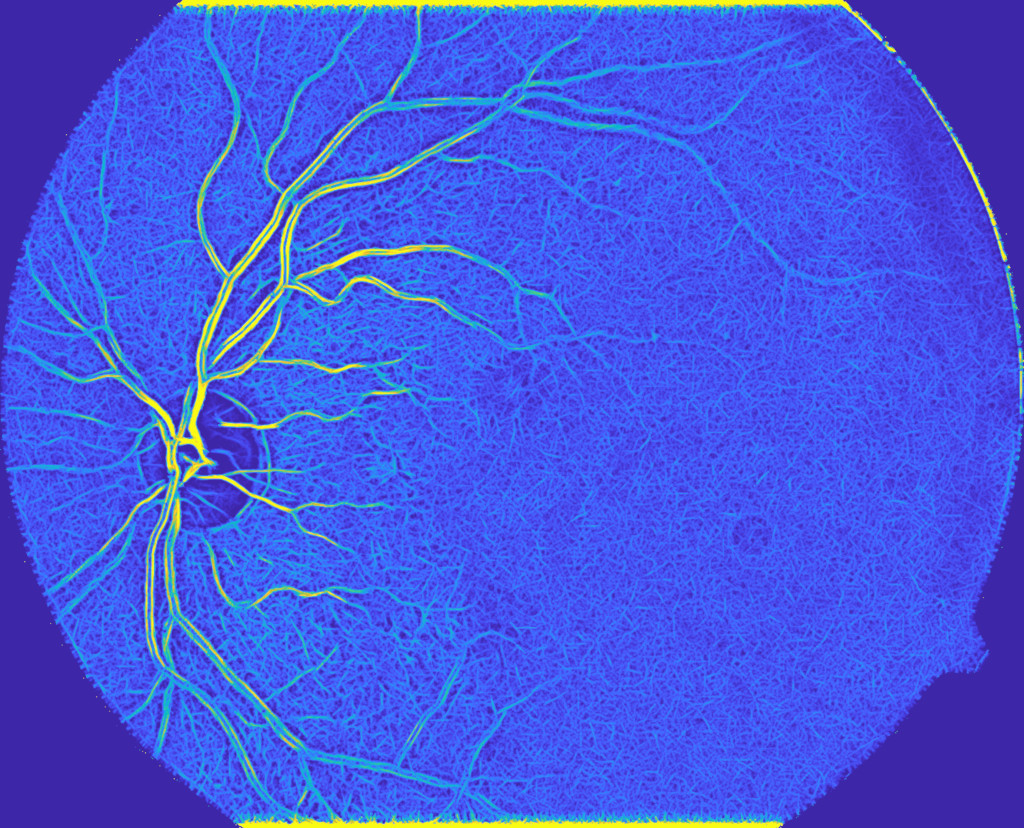}}
  \centerline{(e) B-COSFIRE}\medskip
\end{minipage}
\begin{minipage}[b]{0.32\linewidth}
  \centering
  \centerline{\includegraphics[width=1\columnwidth]{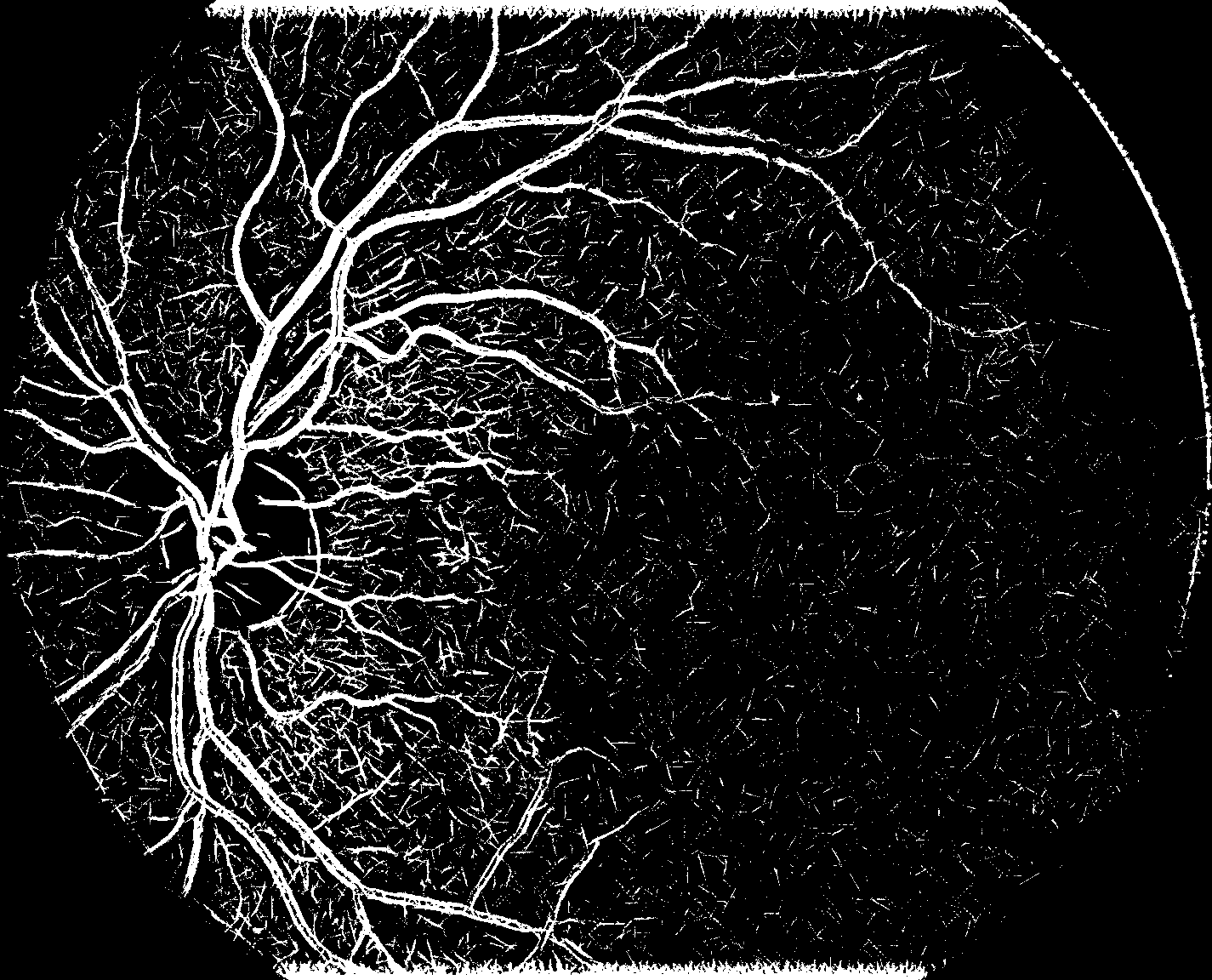}}
  \centerline{(f) Seg. of (e)}\medskip
\end{minipage}
\end{center}
\caption{(a) Map of vesselness $e^I_{b,k} f$. (b) Histogram of $e^I_{b,k} f$ and threshold value in red. (c) Vessel segmentation.  (d) Normalised vesselness $\Phi_{b,k} f$. (e) B-COSFIRE filtered image. (e) Segmentation of (e).}
\label{fig:seg_drk_im}
\end{figure}

The map of vesselness $e^I_{b,k} f$ (Fig. \ref{fig:seg_drk_im} a) is computed for the image of Fig. \ref{fig:profile} (a). 2 probes are automatically selected. 
One can notice that the threshold ($12 \%$ of the FOV area) is below the median (Fig. \ref{fig:seg_drk_im} b). The segmentation (Fig. \ref{fig:seg_drk_im} c) is visually good and allows to detect vessels which are barely visible in the original image (Fig. \ref{fig:profile} a). The normalised map $\Phi_{b,k} f$ (Fig. \ref{fig:seg_drk_im} d) is compared to the vessel detector B-COSFIRE \cite{Azzopardi2015} (Fig. \ref{fig:seg_drk_im} e) whose code is publicly available. In the brightest parts of the image, the B-COSFIRE filter is very efficient to find the vessels and gives more details than our method. However, in the darkest parts, compared to our method, the B-COSFIRE filter is more sensitive to noise and enhances less the vessels. Using the same area threshold, its segmentation detects a lot of noise in addition to the vessels (Fig. \ref{fig:seg_drk_im} f).

\subsection{Quantitative results in a highly contrasted image database}
\label{ssec:exp:drive}

\begin{figure}[htb]
\begin{center}
\begin{minipage}[b]{0.32\linewidth}
  \centering
  \centerline{\includegraphics[width=1\columnwidth]{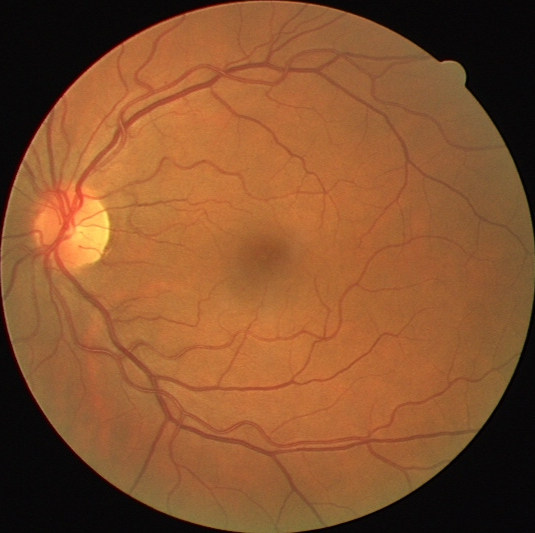}}
  \centerline{(a) Image 1 $f_1$}\medskip
\end{minipage}
\begin{minipage}[b]{0.32\linewidth}
  \centering
  \centerline{\includegraphics[width=1\columnwidth]{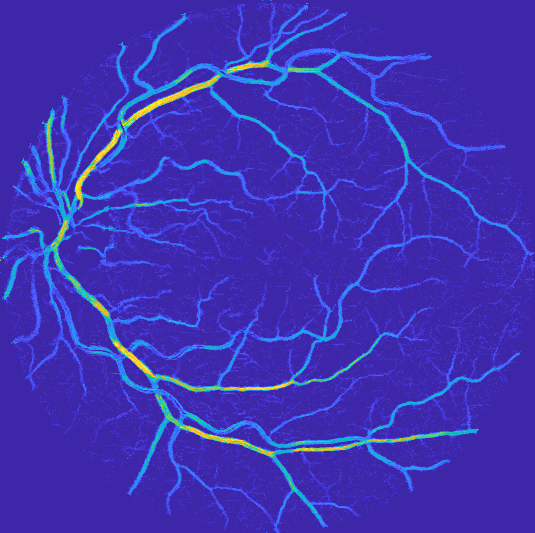}}
  \centerline{(b) $\Phi_{b,k} f_1$}\medskip
\end{minipage}
\begin{minipage}[b]{0.32\linewidth}
  \centering
  \centerline{\includegraphics[width=1\columnwidth]{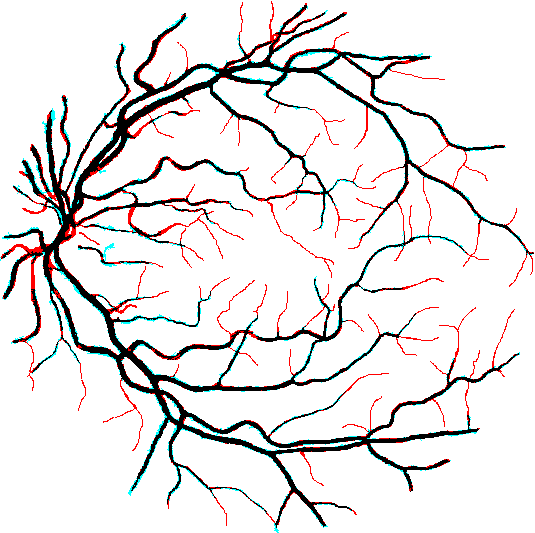}}
  \centerline{(c) Segmentation}\medskip
\end{minipage}\\
\begin{minipage}[b]{0.32\linewidth}
  \centering
  \centerline{\includegraphics[width=1\columnwidth]{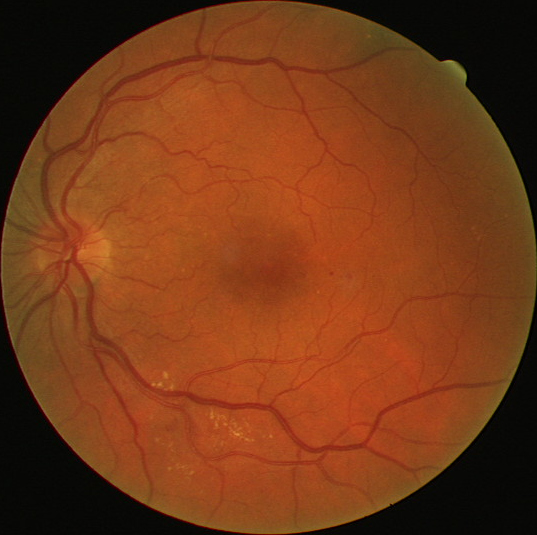}}
  \centerline{(d) Image 3 $f_3$}\medskip
\end{minipage}
\begin{minipage}[b]{0.32\linewidth}
  \centering
  \centerline{\includegraphics[width=1\columnwidth]{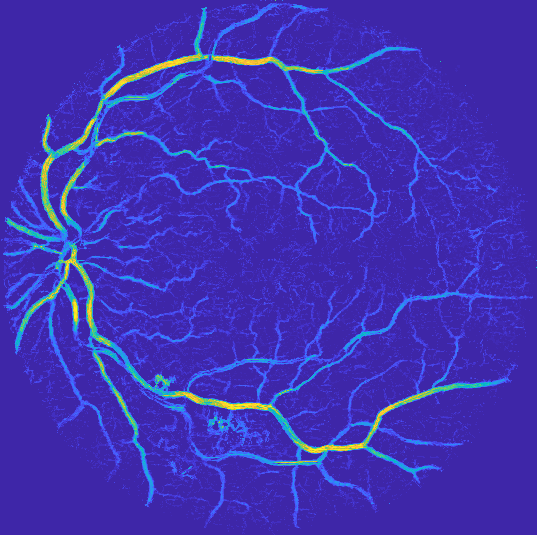}}
  \centerline{(e) $\Phi_{b,k} f_2$}\medskip
\end{minipage}
\begin{minipage}[b]{0.32\linewidth}
  \centering
  \centerline{\includegraphics[width=1\columnwidth]{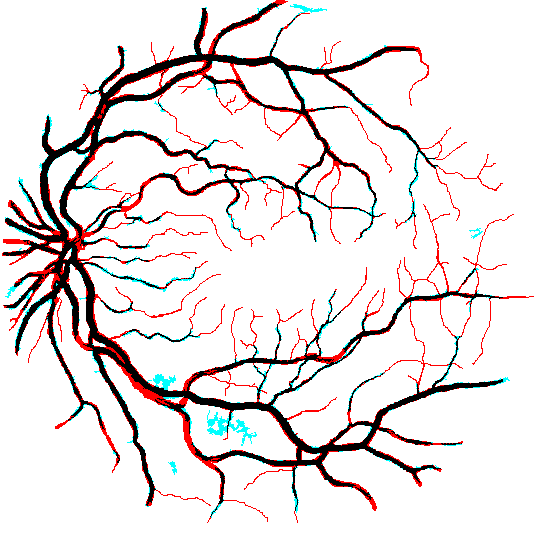}}
  \centerline{(f) Segmentation}\medskip
\end{minipage}
\end{center}
\caption{(a,d) Retinal images. (b,e) Normalised maps. (c,f) Segmentation comparison with the reference. Black pixels are true positives, white pixels are true negatives, cyan pixels are false positives and red pixels are false negatives.}
\label{fig:seg_DRIVE}
\end{figure}

In DRIVE database, as a reference is available we compare the results of our method to those of the $2^{nd}$ expert segmentation (given with the database) and to those of six state-of-the-art methods \cite{Staal2004,Mendonca2006,Azzopardi2015,Zhao2015,Zhu2017,Hu2018}. 
(Tab. \ref{tab:res}). We use the following averaged criteria over the database: the sensitivity (Se), specificity (Sp) and accuracy (Acc) \cite{Staal2004}. 
Using the accuracy criterion, ours is fifth over seven automatic methods. However, when taking into account the standard deviation: ours, the $2^{nd}$ expert and the methods \cite{Mendonca2006,Azzopardi2015,Staal2004} are in the same confidence interval. Three methods  \cite{Zhu2017,Zhao2015,Hu2018} are above the others and the $2^{nd}$ expert. In two images, Fig. \ref{fig:seg_DRIVE} shows that our method is good to find the main vessels (Fig. \ref{fig:seg_DRIVE} c, f). However, it is still limited to segment the smallest ones. In the lower part of Fig. \ref{fig:seg_DRIVE} (f), one can notice that retinopathy lesions such as exudates create false positives (in cyan). Indeed, a thin zone between two exudates can be confounded with a vessel. This will be improved in future works. However, these preliminary results are encouraging because our method is standalone without any pre-processing such as contrast enhancement \cite{Noyel2017c}.

\begin{table}[!htb]
\begin{center}
	\begin{tabular}{@{}l@{ }lll}
	\hline
	Method 																& Se 		 & Sp		  & Acc (std) 	     	 	\\		
	\hline
	Zhu \cite{Zhu2017}										& 0.7140 & 0.9868 & 0.9607 (0.0040)			\\
	Zhao \cite{Zhao2015}									& 0.742	 & 0.982  & 0.954\hspace{0.5em} (-)					\\
	Hu 	\cite{Hu2018}											& 0.7772 & 0.9793 &	0.9533 (-)					\\
	$2^{nd}$ expert	      								&	0.7760 & 0.9725 & 0.9473 (0.0048) 		\\
	Mendon\c{c}a \cite{Mendonca2006}			&	0.7344 & 0.9764 & 0.9463 (0.0065)			\\
	Ours 																	& 0.7358 & 0.9765	& 0.9454 (0.0060)	 		\\
	Azzopardi \cite{Azzopardi2015}				& 0.7655 & 0.9704 & 0.9442 (-)		  		\\
	Staal	\cite{Staal2004}								&	-			 & -			& 0.9441 (0.0065)			\\
	\hline
	\end{tabular}
\end{center}
	\caption{Mean sensitivity (Se), specificity (Sp), accuracy (Acc) and its standard deviation (std) for different methods in DRIVE database.}
	\label{tab:res}
\end{table}

%
%

\section{Conclusion and perpsectives}
\label{sec:concl}

We have successfully introduced a fully automatic method to extract vessels in colour retinal images which is adaptive to lighting variations. It is based on probing from below of an image by a 3-segment probe. A LIP-difference is then locally measured between the image and the probe. This gives a map of vesselness where vessels can be extracted by a threshold. In a lowly contrasted image, results have shown that our method better detects the vessels than a state-of-the-art one \cite{Azzopardi2015}. In a highly contrasted image database (DRIVE), ours gives similar or better results than 3 state-of-the-art ones \cite{Staal2004,Mendonca2006,Azzopardi2015} and the manual segmentation of a second expert. Three methods \cite{Zhu2017,Zhao2015,Hu2018} are above the others and the second expert.
In future, we will make our method more robust to lesions and we will relate it to Mathematical Morphology.

%

\bibliographystyle{IEEEbib}
\bibliography{refs}

\end{document}